\documentclass[conference]{IEEEtran}
\IEEEoverridecommandlockouts
\usepackage[T1]{fontenc}
\usepackage{times}
\usepackage{microtype}
\usepackage{algorithmic}
\usepackage{algorithm}
\usepackage{array}
\usepackage{amsmath,amssymb,amsfonts}
\usepackage[caption=false,font=normalsize,labelfont=sf,textfont=sf]{subfig}
\usepackage{cite}

\usepackage{textcomp}

\usepackage{stfloats}
\usepackage{url}
\usepackage{verbatim}
\usepackage{graphicx}
\usepackage{xcolor}
\usepackage{makecell}
\usepackage{multirow}
\usepackage{amsmath}
\usepackage{fancyhdr}
\usepackage{color}

\usepackage[marginal]{footmisc}
\usepackage{type1cm}

\usepackage[colorlinks,
            linkcolor=blue,
            anchorcolor=blue,
            citecolor=blue,
            urlcolor=blue
            ]{hyperref}

\begin{document}

\title{Automated Heuristic Design for Unit Commitment Using Large Language Models\\
}

\author{
\IEEEauthorblockN{ Junjin Lv}
\IEEEauthorblockA{\textit{Shanghai University of Electric Power} \\
Shanghai, China \\
lvjunjin@mail.shiep.edu.cn}
\\
\IEEEauthorblockN{ Shaodi Zhang}
\IEEEauthorblockA{\textit{Shanghai Electrical Appliances Research Institute (Group) Co} \\
Shanghai, China \\
17321033713@163.com}
\\
\IEEEauthorblockN{ Chunyang Gong}
\IEEEauthorblockA{\textit{Shanghai University of Electric Power} \\
Shanghai, China \\
gongchunyang@shiep.edu.cn}
\and
\IEEEauthorblockN{  Chenggang Cui}
\IEEEauthorblockA{\textit{Shanghai University of Electric Power} \\
Shanghai, China \\
cgcui@shiep.edu.cn}
\\
\IEEEauthorblockN{ Hui Chen}
\IEEEauthorblockA{\textit{Shanghai University of Electric Power} \\
Shanghai, China \\
chenhui@shiep.edu.cn}
\\
\IEEEauthorblockN{ Jiaming Liu}
\IEEEauthorblockA{\textit{Shanghai University of Electric Power} \\
Shanghai, China \\
y23204141@mail.shiep.edu.cn}
}

\maketitle
\thispagestyle{plain}
\renewcommand{\thefootnote}
\footnote{}
\renewcommand{\footrulewidth}{1pt}

\begin{abstract}
The Unit Commitment (UC) problem is a classic challenge in the optimal scheduling of power systems. Years of research and practice have shown that formulating reasonable unit commitment plans can significantly improve the economic efficiency of power systems' operations. In recent years, with the introduction of technologies such as machine learning and the Lagrangian relaxation method, the solution methods for the UC problem have become increasingly diversified, but still face challenges in terms of accuracy and robustness. This paper proposes a Function Space Search (FunSearch) method based on large language models. This method combines pre-trained large language models and evaluators to creatively generate solutions through the program search and evolution process while ensuring their rationality. In simulation experiments, a case of unit commitment with \(10\) units is used mainly. Compared to the genetic algorithm, the results show that FunSearch performs better in terms of sampling time, evaluation time, and total operating cost of the system, demonstrating its great potential as an effective tool for solving the UC problem.
\end{abstract}

\begin{IEEEkeywords}
Large Language Model, Unit Combination, Electric Power System, Function Space Search, Operating Cost
\end{IEEEkeywords}

\section{Introduction}
The unit commitment problem (UC) is a classic issue in optimal power system scheduling. Years of theoretical research and practical operation have shown that formulating a reasonable and optimized unit commitment plan can significantly improve the economic operation efficiency of the power system \cite{b1}.

In recent years, numerous scholars have conducted in-depth studies on the unit commitment (UC) problem and proposed various optimization methods to solve it. For example, Reference \cite{b2} proposes a UC model that combines machine learning and Lagrangian relaxation methods, improving the efficiency of the solution of complex UC problems through online self-learning; Reference \cite{b3} studies the multiobjective UC problem based on profit maximization, comprehensively considering the uncertainties of renewable energy sources and energy storage units; Reference \cite{b4} proposes an algorithm that combines discrete differential evolution and quadratic programming to optimize the start-stop status of units and economic dispatch in a layered manner. Compared with traditional optimization methods, the above-mentioned methods have relatively obvious effects in solving the UC problem. However, there are still certain limitations in terms of accuracy and robustness in solving the unit commitment problem.

Recently, with the rise of Large Language Model (LLM) technology, a revolution is quietly unfolding across multiple industries. In the field of power systems, the combination of LLM with advanced optimization algorithms and data-driven strategies is injecting strong momentum into the evolution of smart grids, ushering in unprecedented development opportunities. For example, Reference \cite{b5} analyzes the potential security threats that large language models may bring to modern power systems, emphasizing the necessity of researching and developing countermeasures; Reference \cite{b6} discusses the potential of foundation models (such as GPT-4) in power system tasks and evaluates their performance in tasks such as optimal power flow, vehicle scheduling, and technical report knowledge retrieval; Reference \cite{b7} proposes a large language model based on prompt templates to generate power grid operation instructions and improves the quality of generation through human expert feedback. LLM has shown great potential in solving complex tasks. However, LLM sometimes suffers from the "hallucination" problem \cite{b8}, which may lead to seemingly reasonable but incorrect explanations.

In view of the existing problems in current academic research, this study introduces a Function Space Search (FunSearch) method. This is an evolutionary process that pairs a pre-trained LLM with a system evaluator. It combines large language models and program search to solve complex mathematical and computational problems. This method performs excellently in solving the unit commitment problem in power systems and can efficiently address related challenges and optimize the scheduling and operation of generating units.

\section{Large Language Model UC Decision Optimization Framework}

As shown in Fig.\ref{Large Language Model UC Optimization Framework}, the Large Language Model Unit Commitment (LLM UC) Decision Optimization Framework is shown in Figure 1. In the LLM UC Decision Optimization Framework, key components include generating units, load demand dispatch centers, LLM inference modules, and FunSearch optimization modules. This framework ingeniously combines pre-trained LLMs with evaluators. It can not only understand complex power system operation rules but also intelligently generate and continuously optimize generator scheduling plans according to real-time data. Through the iterative evolution process, the LLM UC framework can quickly find scheduling strategies close to the optimal solution, effectively reducing the operating costs of the power system and improving economic benefits.
\begin{figure}[htbp]
\vspace{-1.0em}
    \centering
    \includegraphics[width=1\linewidth]{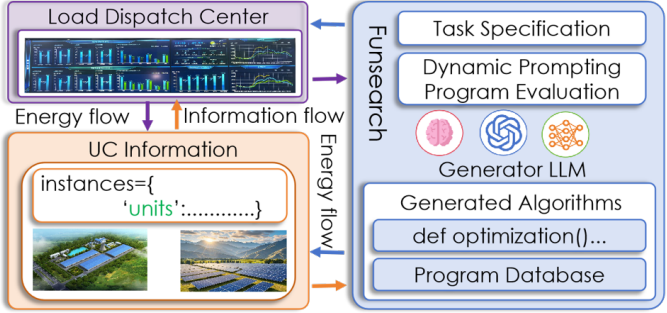}
    \caption{{Large Language Model UC Optimization Framework}}
    \label{Large_Language_Model_UC_Optimization_Framework}
\vspace{-1.0em}
\end{figure}

\section{Mathematical model of unit commitment}

The UC problem is a classical optimal scheduling problem in power systems, where the main objective is to schedule a group of generating units in a given time period so that the operating cost of the whole system is minimized and the load demand, as well as various constraints are satisfied at the same time.

\subsection{objective function}\label{AA}
The UC problem is to rationally determine the optimal startup and shutdown states and the corresponding output sizes of all available units in each computation period to minimize the total system operating cost, subject to various constraints.The UC problem is a large-scale mixed integer nonlinear constrained optimization problem, which contains the \(0\) and \(1\) decision variables that represent the startup and shutdown states of the units, and a continuous variable for the output power. The total system operating cost consists of two parts: unit operating cost and penalty cost, and its objective function can be expressed as:
\begin{equation}
C_{\mathrm{total}} = C_o + P_d + P_{\min_t}
\end{equation}

\begin{equation}
C_o = \sum_{t=1}^T \sum_{i=1}^N C_i(u_i^t) \cdot P_i^t
\end{equation}

\begin{equation}
P_d = 10^4 \sum_{t=1}^T \max \left(0, D_t - \sum_{i=1}^N u_i^t \cdot P_i^t \right)
\end{equation}

\begin{equation}
P_{\min_t} = \sum_{i=1}^N \sum_{j=1}^{M_i} 
\begin{cases} 
10^5 (UT_i - l_{i,j}), & \text{if } l_{i,j} < UT_i \\
0, & \text{otherwise}
\end{cases}
\end{equation}

where \( C_{\mathrm{total}} \), \( C_o \), \( P_d \), \( P_{\min_t} \) denote the total system operating cost, the unit operating cost, the penalty cost when demand is not met, and the penalty cost for the minimum start/stop time constraint, respectively; \( C_i(u_i^t) \) is the cost function of the generating unit \( i \) at time period \( t \); \( u_i^t \) is the start/stop state of the unit (1 or 0); \( P_i^t \) is the power generation of the unit \( i \) at time period \( t \); \( D_t \) is the demand load at time period \( t \); \( UT_i \) is the minimum start/stop time of the unit \( i \); and \( l_{i,j} \) is the time period of the unit \( i \) at the first \( j \) length of continuous operation or shutdown.

\subsection{Restrictive Condition}
\subsubsection{Load Balancing Constraint}For each time period \(t\) the total generation of the system must meet the power demand for that time period.
\begin{equation}
\sum\limits_{i = 1}^N {\mathop P\nolimits_i^t }  = \mathop D\nolimits_t ,\forall t \in \left\{ {1,2, \cdot  \cdot  \cdot T} \right\}
\end{equation}

\subsubsection{Unit Start-Stop Constraint}Each unit \(i\) can only be in the on or off state at each time period \(t\).
\begin{equation}
u_i^t \in \{0,1\}, \quad 
\begin{cases} 
\forall t \in \{1,2,\dots,T\} \\
\forall i \in \{1,2,\dots,N\}
\end{cases}
\end{equation}

where \(\mathop u\nolimits_i^t  = 1\) indicates the on state and \(\mathop u\nolimits_i^t  = 0\) indicates the off state.

\subsubsection{Minimum Start-Stop Time Constraint}Each unit needs to be maintained for a certain period of time after startup or shutdown before switching states, and the constraints are as follows for the minimum run time after startup \(\mathop T\nolimits_{\min \_up}\) and the minimum shutdown time after shutdown\(\mathop T\nolimits_{\min\_down}\).

Minimum runtime after power-up:Indicates that the unit's continuous operation time after startup cannot be less than the minimum startup time.
\begin{equation}
\sum\limits_{\mathop t\nolimits^ *   = t}^{t + \mathop T\nolimits_{\min_up,i} } {\mathop u\nolimits_i^{\mathop t\nolimits^ *  } }  \geqslant \mathop T\nolimits_{\min_up,i} 
\end{equation}

Minimum running time after shutdown: indicates that the unit's continuous downtime after shutdown cannot be less than the minimum downtime.
\begin{equation}
\sum\limits_{\mathop t\nolimits^ * = t}^{t + \mathop T\nolimits_{\min \_down,i} } {(1 - \mathop u\nolimits_i^{\mathop t\nolimits^ *  } )}  \geqslant \mathop T\nolimits_{\min \_down,i}
\end{equation}

\subsubsection{Upper and Lower Limits of Unit Output Constraints}
\begin{equation}
\mathop P\nolimits_{i,\min }^t \leqslant \mathop P\nolimits_i^t  \leqslant \mathop P\nolimits_{i,\max }^t
\end{equation}

\( P_{i,\min}^t \) and \( P_{i,\max}^t \) indicate the upper and lower limits of Unit \( i \) output during the time period \( t \).

\thispagestyle{plain} 

\section{ The Evolution of FunSearch Program Search }
\subsection{FunSearch Methods}

FunSearch is a search method that combines a pre-trained LLM and an evaluator designed to generate solutions creatively while ensuring their reasonableness \cite{b8}. A UC decision scheme is generated by calling the LLM on the content and initial code in the prompt, and the evaluator reviews this output, filtering and improving the unreasonable parts. The two work in tandem through an iterative process: the LLM generates scenarios and the evaluator evaluates and provides feedback, leading to incremental improvements in the quality of the scenarios.

\subsection{Program Search Evolution}
Program Search is a method that uses computer programs to create and improve solutions. It combines search algorithms with program generation methods and keeps exploring new solutions to optimize problem solving.  

In the FunSearch method, Program Search is used to continue evolving the initial program in order to lower system operating costs and solve the unit commitment problem. The evolution of FunSearch is shown in Fig.\ref{fig:FunSearch-Evolution-Process}.

\begin{figure*}[htbp]
    \vspace{-1.0em}
    \centering
    \includegraphics[width=0.8\textwidth]{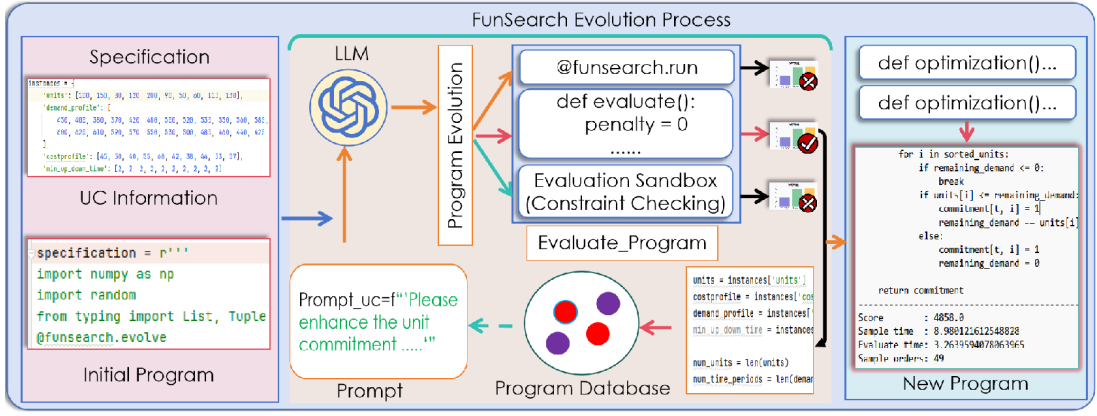}
    \caption{{FunSearch Evolution Process}} 
    \label{fig:FunSearch_Evolution_Process}
    \vspace{-1.0em}
\end{figure*}

\subsubsection{Input}In FunSearch, the input consists primarily of unit information and an initial program. Unit information includes unit power, load demand, minimum startup/shutdown times, and cost profiles. A simple initial program is provided for further evolution.

\subsubsection{Prompt Engineering}By using the input information, the packing program database, and the need for program improvements, we feed these elements into the LLM to generate new programs.

\subsubsection{Invoke the LLM API}In FunSearch, the LLM serves as the core. It uses the given prompts to generate new programs that meet the required needs.

\subsubsection{Program Evaluation}Using the evaluation program, we assess the new programs generated by the LLM in terms of their score (total unit operating cost), evaluation time, and sampling time. Programs with a valid score are then placed in the program database for packing, while those without a score or exceeding the time limit are discarded.

\subsubsection{Program Database}After evaluation, the newly developed heuristics are compiled into a library of programs awaiting iterative comparisons with subsequently evaluated heuristic implementations. The aggregated insights and performance metrics are then systematically integrated into the generation of hints, enabling the LLM to use this contextual knowledge as a reference to synthesize new heuristics in subsequent optimization cycles.

\subsubsection{Output}After comparison, the main output is the new program with the lowest UC operating cost score and the on/off decisions for each unit at every time step. The FunSearch solving process is shown in Fig.\ref{FunSearch Solving Process}.
\begin{figure}[htbp]
\vspace{-1.0em}
    \centering
    \includegraphics[width=1\linewidth]{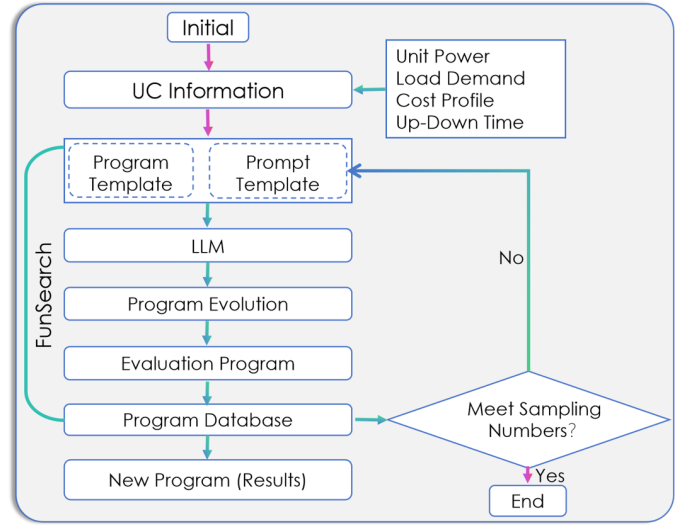}
    \caption{{FunSearch Solving Process}}
    \label{FunSearch_Solving_Process}
\vspace{-1.0em}
\end{figure}

\thispagestyle{plain} 
\section{Case Study}
To verify the effectiveness of the proposed FunSearch method, we programmed in Python \(3.9\) and tested it on a Chengde system that includes \(10\) units. This system divides one day into \(24\) time periods, treating each hour as a separate calculation interval. Using data for 10 units and 24 periods provided in \cite{b9}, we ran the algorithm 50 consecutive times and performed a statistical analysis of the results. Finally, we chose the best solution from these \(20\) runs as the final optimized result.

To evaluate the effectiveness of FunSearch in determining the initial solution for the UC problem, we compared the solution of the UC problem with a Genetic Algorithm (GA) versus using FunSearch. We mainly compared the sampling time, the evaluation time, and the total operating cost of the system. The comparison results are shown in Table \ref{Comparison of the Results from the Two Methods}.
\begin{table}[htbp]
\vspace{-1.0em}
        \centering
        \caption{Comparison Results Analysis} 
        \label{Comparison of the Results from the Two Methods}
        \begin{tabular}{c| c |c| c}
                \hline\hline\noalign{\smallskip}        
\multirow{2}*{Approach}&
           {Sampling Time} &{Evaluation Time}&{Operating Cost}  \\
           &{(s)}&{ (s)}& {(dollar(s)}\\
             \noalign{\smallskip}\hline\noalign{\smallskip}
            GA & 240 & 14.5 & 5236 {\smallskip}\\
            FunSearch & 6.6 & 3.9 & 4884 {\smallskip}\\
        \hline\hline
        \end{tabular}
\end{table}

From Table \ref{Comparison of the Results from the Two Methods}, we see that compared to the GA method, FunSearch performs better in both sampling time and evaluation time, and the optimal schedule it produces has lower operating costs.

The visualization results and the on/off decisions of each unit when solving the UC problem with the Genetic Algorithm are shown in Fig.\ref{Visualization of the GA Solution for the UC Problem}.
\begin{figure}[htbp]
\vspace{-1.0em}
    \centering
    \includegraphics[width=1\linewidth]{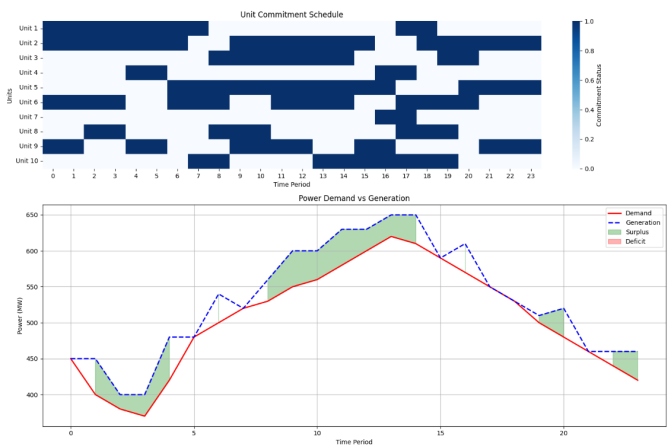}
    \caption{Visualization of the GA Solution for the UC Problem}
    \label{Visualization_of_the_GA_Solution_for_the_UC_Problem}
\vspace{-1.0em}
\end{figure}

The visualization results and the on/off decisions of each unit when solving the UC problem with FunSearch are shown in Fig.\ref{Visualization of the FunSearch Solution for the UC Problem}.
\begin{figure}[htbp]
\vspace{-1.0em}
    \centering
    \includegraphics[width=1\linewidth]{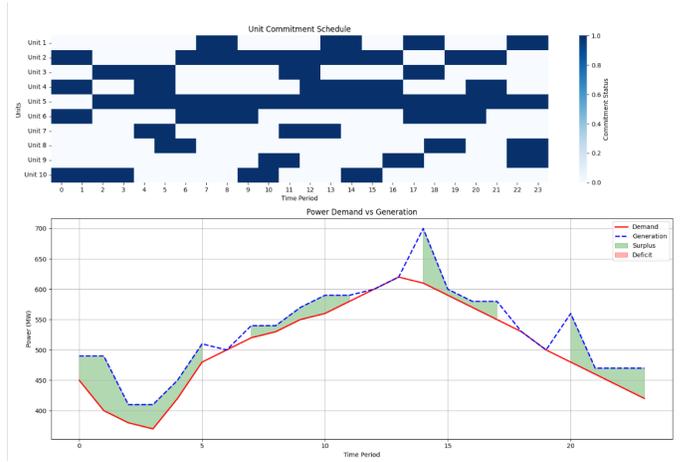}
    \caption{Visualization of the FunSearch Solution for the UC Problem}
    \label{Visualization_of_the_FunSearch_Solution_for_the_UC_Problem}
\vspace{-1.0em}
\end{figure}
Based on the overall results of the visual comparison, we can see that in most cases, FunSearch performs better than the GA algorithm in generating an initial solution for the UC problem, and in certain situations, it can successfully find the optimal solution. This shows the strong potential of FunSearch as an effective tool for solving the UC problem.

\section{conclusion}
This research explores the innovative application of Large Language Models (LLM) in optimizing complex power system Unit Commitment (UC) decisions. By proposing a Function Space Search (FunSearch) method that combines pre-trained language models with a system evaluator, the study aims to solve accuracy and robustness challenges in traditional UC problem solving.
The method leverages LLM's semantic understanding and program search techniques, iteratively generating and precisely evaluating programs to intelligently optimize power generator scheduling and effectively reduce system operating costs. The simulation results show that, compared to traditional genetic algorithms, FunSearch performs exceptionally well in solving efficiency, computation time, and cost control.
Future research can further improve the model's reliability and generalization capabilities by addressing potential "hallucination" issues when LLM handles UC problems.

\vspace{10pt}

\thispagestyle{plain} 
\end{document}